\pdfoutput=1
\documentclass[11pt]{article}
\usepackage[dvipsnames]{xcolor}
\usepackage{acl}%

\usepackage{hyperref}
\usepackage{url}

\usepackage{times}
\usepackage{amsmath}
\usepackage{times}
\usepackage{graphicx}
\usepackage{latexsym}
\usepackage[capitalize]{cleveref}
\usepackage{microtype}
\usepackage[T1]{fontenc}
\usepackage[utf8]{inputenc}
\usepackage{tabularx}
\usepackage{booktabs}
\usepackage{multirow}
\usepackage{xspace}
\usepackage{floatrow}
\usepackage{enumitem}
\setlist[enumerate]{itemsep=0mm}
\usepackage{fontawesome} %
\usepackage{amssymb}%
\usepackage{pifont}%

\graphicspath{{images/}}
\usepackage{subfig}
\usepackage{worldflags}

\setkeys{Gin}{width=\textwidth}
\newcommand{\xmark}{\ding{55}}%
\newcolumntype{P}[1]{>{\centering\arraybackslash}p{#1}}
\newcolumntype{M}[1]{>{\centering\arraybackslash}m{#1}}

\makeatletter
\g@addto@macro\@floatboxreset\centering
\makeatother

\usepackage{xspace}

\makeatletter
\DeclareRobustCommand\onedot{\futurelet\@let@token\@onedot}
\def\@onedot{\ifx\@let@token.\else.\null\fi\xspace}

\makeatother

\title{Cross-cultural Inspiration Detection and Analysis \\ in Real and LLM-generated Social Media Data}

 \author{Oana Ignat$^1$\thanks{Oana Ignat and Gayathri Ganesh Lakshmy contributed equally to the manuscript} \hspace{5pt} 
 Gayathri Ganesh Lakshmy$^2{^*}$ \hspace{5pt}  
Rada Mihalcea$^1$ \\
$^1$University of Michigan - Ann Arbor, USA \\
$^2$Sri Sivasubramaniya Nadar College of Engineering - Chennai, Tamil Nadu, India \\
\textit{\{oignat, mihalcea\}@umich.edu} \hspace{5pt}  \textit{gayathri2010090@ssn.edu.in}\\ }

\begin{document}

\maketitle

\begin{abstract}
Inspiration is linked to various positive outcomes, such as increased creativity, productivity, and happiness. Although inspiration has great potential, there has been limited effort toward identifying content that is {\it inspiring}, as opposed to just engaging or positive. Additionally, most research has concentrated on Western data, with little attention paid to other cultures. 
This work is the first to study cross-cultural inspiration through machine learning methods. 
We aim to identify and analyze real and AI-generated cross-cultural inspiring posts. To this end, we make publicly available the \textsc{InspAIred} dataset, which consists of 2,000 real inspiring posts, 2,000 real non-inspiring posts, and 2,000 generated inspiring posts evenly distributed across India and the UK. The real posts are sourced from Reddit, while the generated posts are created using the GPT-4 model. 
Using this dataset, we conduct extensive computational linguistic analyses to (1) compare inspiring content across cultures, (2) compare AI-generated inspiring posts to real inspiring posts, and (3) determine if detection models can accurately distinguish between inspiring content across cultures and data sources.
Our dataset can be accessed alongside our classification models at \url{https://github.com/MichiganNLP/cross_inspiration}. 
\end{abstract}

\section{Introduction}
Inspiration has been a part of our world for millennia, starting with ancient Greece, where Muses were responsible for delivering divine knowledge by whispering in a poet's ear~\cite{leavitt1997poetry}, all the way to today's creativity domain, where it is still common for artists and scientists to attribute their best ideas to a higher power, independent of their own control.

According to \citet{Thrash2003InspirationAA, Thrash2004InspirationCC}, inspiration is a general construct that consists of three core characteristics: \textit{evocation}, \textit{transcendence}, and \textit{approach motivation}. Evocation refers to the process of being triggered by a stimulus, either from within (such as a creative idea that comes from the subconscious) or from outside (such as a person, object, music, or nature). Transcendence allows one to perceive something beyond their usual concerns~\citep{Milyavskaya2012InspiredTG}. Finally, an inspired person is motivated to express, transmit, or \textit{act} on their inspiration~\citep{elliot2002approach}.

Inspiration is an area of study with promising cross-disciplinary applications in creative fields (e.g., advertisement, storytelling), education, therapy, mentorship, coaching, or social media.
For instance, social network recommendation systems can mitigate potential harms by showing more positive and inspiring content to users~\citep{Ignat2021DetectingIC}. 
Access to inspiring content can have a positive impact on people's lives by offering them a fresh perspective and motivating them to take action, particularly during periods of uncertainty and concern~\citep{Oleynick2014TheSS}. 
Moreover, inspiration facilitates progress towards goals~\cite{Milyavskaya2012InspiredTG} and increases overall well-being~\cite{Thrash2010MediatingBT}. 

\begin{figure}
\centering
\includegraphics[width=\linewidth]{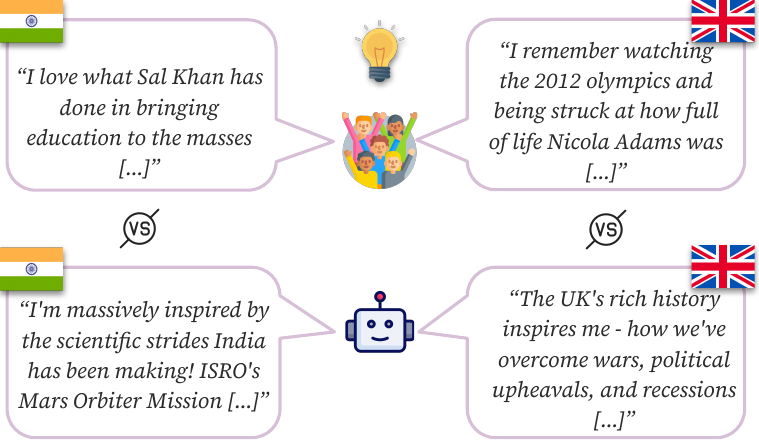}
\caption{We compare AI-generated and human-written inspiring Reddit content across India and the UK. Although it is challenging for a person to distinguish between them, we find significant linguistic cross-cultural differences between generated and real inspiring posts.\vspace{-2em}}
\label{fig:main_idea}
\end{figure}

Despite the compelling motivation, little research has been done on the automatic identification of content that is {\it inspiring}, as opposed to simply positive~\cite{Ignat2021DetectingIC}, and no studies have been conducted on how inspiration varies across cultures, and whether it can be automatically generated.
At the same time, LLMs' impressive generative ability can create new opportunities for automatically generating inspiring content.

In this work, we aim to address these research gaps by focusing on three main research questions. 
(RQ1:) How do inspiring posts compare across cultures? (RQ2:) How do AI-generated inspiring posts compare to real inspiring posts across cultures? and (RQ3:) Can detection models differentiate inspiring posts across cultures and data sources?

We summarize our contributions as follows. 
First, we share {\bf a novel dataset comprised of cross-cultural real and generated inspiring and non-inspiring posts}, for a total of 2,000 real inspiring posts, 2,000 non-inspiring posts, and 2,000 LLM-generated posts, balanced across India and the UK. 
Second, we make use of the dataset to conduct {\bf computational linguistic analyses to compare inspiring content across cultures and data sources}, i.e., India vs. the UK and AI-generated vs. human-written. Finally, {\bf we explore the effectiveness of machine learning models for detecting inspiration} across diverse data sources and cultures.

\section{Related Work}

\paragraph{Automatic Inspiration Detection.}
There have been a limited number of research studies conducted on inspiration, with the majority of them being carried out by the psychology and sociology communities.
These studies, such as the ones by \citet{Thrash2003InspirationAA, Thrash2004InspirationCC} and \citet{elliot2002approach}, have established the fundamental characteristics of inspiration. Additionally, they have developed a scale to measure the frequency with which people experience inspiration in their lives. These studies found that individuals who are inspired tend to be more open to new experiences and show greater absorption in their tasks. They are also more intrinsically motivated, have a strong drive to master their work, and are less competitive.

The work most similar to ours is from \citet{Ignat2021DetectingIC}, who are the first to study inspiration through machine learning methods. 
To facilitate research in this domain, they release a weakly labeled dataset consisting of inspiring and non-inspiring posts collected from Reddit and annotate these posts with their effect on the reader and the emotions they transmit. 
They also provide a RoBERTa~\citep{Liu2019RoBERTaAR} classifier fine-tuned on human labels to provide a strong baseline for determining whether a post is inspiring. 
Finally, they perform data analyses to gain insights into which topics inspire readers and how they influence them.
Our work builds on \citet{Ignat2021DetectingIC} by extending it to other cultures, India and the UK, and by collecting AI-generated inspiring posts in order to compare them to human-written posts. 

\paragraph{Human vs. LLM-generated Cross-cultural Text.}
With the rapid development of LLMs, these models demonstrate remarkable proficiency in generating human-like text across multiple languages and styles~\cite{Wu2023ASO, tang2023science}. 

In particular, LLMs excel at creative writing, such as story generation~\citep{yuan2022wordcraft}, advertising slogan creation ~\citep{murakami2023natural}, and news composition~\citep{yanagi2020fake}.
Tools like ~\citet{yuan2022wordcraft} can help users in their creative pursuits. By generating inspiring content, our work is more indirect but still aims to help the user express and act on their insights. 

More similar to our work, LLMs have also started to be applied to motivate people~\cite{cox2023prompting}. For example, ~\citet{karinshak2023working} used GPT-3 to generate messages to persuade people to receive the Covid-19 vaccine. To the best of our knowledge, we are the first to generate inspiring content using LLMs.
Our work is also part of the emerging work on modeling cultural factors in LLMs~\citep{huang2023culturally, fung2022normsage, ramezani2023knowledge}.
Inspiration varies across cultures. Therefore, we test the cultural knowledge of LLMs about inspiration in India and the UK, and compare it to inspiring Reddit posts from users in these countries.

\paragraph{Computational Linguistics for Social Media Analysis.}
The advent of computational linguistics techniques has enabled researchers to analyze vast amounts of social media data for various purposes, including sentiment analysis, topic modeling, and linguistic variation across cultures~\citep{Pennebaker2007LinguisticIA, pang2008opinion, imran2020cross}. These works have facilitated the extraction of meaningful insights from diverse linguistic contexts, paving the way for studies on cross-cultural communication in online environments.

Social media is a key source of inspiring content for younger audiences \citep{raney2018profiling}. Features such as hope and appreciation of beauty and excellence trigger self-transcendent emotions in videos tagged with ``inspiration'' on YouTube \citep{dale2017youtube}, as well as in \#inspiring and \#meaningful Tumblr memes and Facebook posts \citep{rieger2019daily, dale2020self}.
Similarly, we collect Reddit posts from subreddits related to inspiration across UK and Indian cultures and analyze them using computational linguistic tools such as \citet{Pennebaker2007LinguisticIA}.

\section{The \textsc{InspAIred} Dataset}
To answer our research questions, we compile a novel dataset, which we refer to as \textsc{InspAIred} - AI-generated Inspiring Reddit Content. 
Our dataset contains inspiring and non-inspiring posts from India and the UK, from two different sources: (1) crawled from Reddit and (2) generated by an LLM.

\subsection{{\color{Thistle}\faUsers} Real Inspiring Content}
We collect 2,000 weakly labeled inspiring posts and 2,000 weakly labeled non-inspiring posts, balanced across India and the UK. We describe our data collection and annotation process below. 

\paragraph{Data Collection.}
We scrape around 5,300 posts from Reddit, a popular online platform, specifically focusing on culturally inspiring content. 
Following the data collection process from~\citet{Ignat2021DetectingIC}, we conduct searches within culture-related or discussion-related flairs of the subreddits using keywords, such as ``inspiration'' and ``motivation'', to identify the relevant data in the form of both posts and comments.

More specifically, for Indian data, we primarily target the regions of \textit{Kerala, Karnataka, Maharashtra}, and \textit{Tamil Nadu}. Besides the general ``r/india'', we also explore subreddits at the \textit{state} level, such as ``r/Kerala'', ``r/Karnataka'', ``r/Maharashtra'', and ``r/TamilNadu'', which serve as hubs for focused discussions on regional culture, traditions, and social issues. Finally, we also examine subreddits from specific \textit{cities}, including capital cities like \textit{Chennai} and \textit{Bangalore}, to capture more local perspectives and experiences.
For the UK data, we follow a similar strategy to collect Reddit posts, targeting both \textit{state} level subreddits, such as ``r/UnitedKingdom'', as well as \textit{regional} subreddits representing areas and capital cities, such as ``r/London''.

We group the collected posts into two main categories: inspiring posts from India and inspiring posts from the UK. Initially, we attempted to create a more fine-grained split at the region or city level, but we faced difficulties in finding annotators from those specific regions. However, we encourage future research to explore this direction and investigate how inspiration varies within a country or within a region, along with exploring other demographic information such as language, age, gender, or income. 

\begin{figure}
\centering
\includegraphics[width=\linewidth]{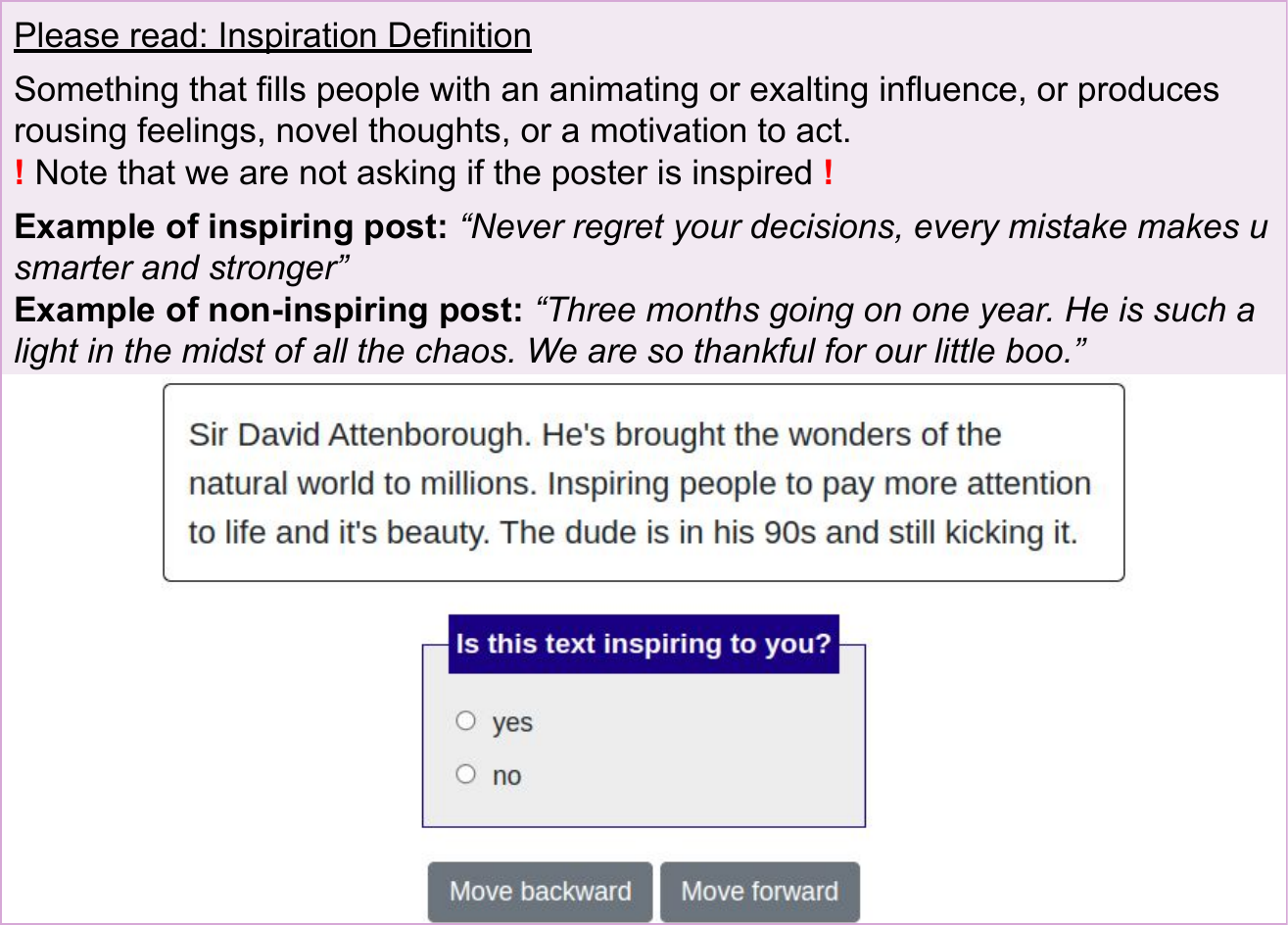}
\caption{Annotation guidelines for labeling inspiration.\vspace{-1.8em}}
\label{fig:UI_annotations}
\end{figure}

\paragraph{Data Filtering.} \label{data-filter}
The collected posts are further filtered with an inspiration classifier model. 
We use the XLM-RoBERTa base~\cite{roberta} from \texttt{HuggingFace}.\footnote{\url{https://huggingface.co/FacebookAI/xlm-roberta-base}}
The model is a multilingual version of RoBERTa~\cite{Liu2019RoBERTaAR}, and is pre-trained on 2.5TB of filtered CommonCrawl~\cite{wenzek-etal-2020-ccnet}\footnote{We choose a multilingual model, as the Indian data contains code-mix data.}. 
We fine-tune the model for five epochs using the dataset described below, with a learning rate of $2e-5$ and a batch size of $8$. More implementation details can be found in the ~\Cref{finetune}.

For fine-tuning, we use the data from \citet{Ignat2021DetectingIC}. 
The data contains around 12,000 annotated ``general'', i.e., no target culture, balanced Reddit inspiring and non-inspiring posts. More details about the fine-tuning data can be found in~\citet{Ignat2021DetectingIC}.
Next, we evaluate the quality of the fine-tuned model predictions by annotating a subset of the posts and comparing them with the model predictions.

\paragraph{Data Annotation.}
The posts are annotated by crowd-sourced workers from India and the UK. For each country, we select a sample of 200/ 1,000 posts equally sampled from inspiring and non-inspiring predicted posts, to be labeled by three annotators from that country.

We use Potato~\cite{pei2022potato} to create the user interface and store the data. 
Following the guidelines from \citet{Ignat2021DetectingIC}, the user interface contains a definition of inspiration and examples of inspiring and non-inspiring posts, as seen in \Cref{fig:UI_annotations}.
To find and hire UK annotators, we connect Potato to Prolific.\footnote{\url{www.prolific.com}} We select three annotators who have the following qualifications: an approval rate >98\%, living in the UK, and a right to vote (as a proxy for age). 
The annotators independently label 200 posts each, categorizing the posts based on their subjective judgment of whether they found the content inspiring or not.
We conduct a similar annotation process with three reliable Indian annotators to label the data from India. 
To ensure consistency amongst the annotators, they are provided with the same interface as used for the UK annotators and shown in \Cref{fig:UI_annotations}. 

Finally, we compute the agreement score between the annotators using the Fleiss Kappa measure \cite{fleiss1973equivalence}. We obtain a score of $0.24$ for UK data and $0.29$ for Indian data, indicating a fair agreement, aligning with   ~\citet{Ignat2021DetectingIC} who obtained a $0.26$ agreement score.
Furthermore, by manually examining the low-agreement examples, we find that inspiration is highly subjective and \textit{personal}, relying on one's beliefs, interests, or upbringing (e.g., someone not interested in nature will probably not be inspired by David Attenborough). Finally, we make available all the annotations to encourage future work to research the general or personal nature of inspiration based on data and downstream applications.

Given that each post is annotated by three annotators, following ~\citet{Ignat2021DetectingIC}, we mark a post as inspiring if at least one annotator labeled it as inspiring. We do not select the majority label to avoid excluding other annotators’ opinions. Disagreement among annotators in subjective tasks may reflect systematic differences in opinions across groups, not noise~\citep{fleisig-etal-2023-majority}.

We find that the predictions made by the fine-tuned model are quite similar to human annotations. For Indian data, the accuracy rate is $72.9\%$, and the F1 score is $75.5\%$. Similarly, for UK data, the accuracy rate is $73.5\%$, and the F1 score is $80.1\%$.
Therefore, we decide to not annotate more data and instead use the model predictions. Additionally, we further fine-tune the model on the annotated data and use it to collect more weakly labeled data. This approach aims to leverage the insights gained from the initial fine-tuning process and apply them to the user-annotated subset, thereby refining the model to better capture the nuances present in this specific dataset. We use the fine-tuned model predictions and the annotated subset as our final labeled data.

\paragraph{Quality Assurance.} 
We remove posts that are classified as toxic or hate speech using two fine-tuned XLM-RoBERTa classifiers from \texttt{HuggingFace}.\footnote{\url{https://huggingface.co/textdetox/xlmr-large-toxicity-classifier}}\footnote{\url{https://huggingface.co/facebook/roberta-hate-speech-dynabench-r4-target}}
We also remove profanity from posts using a profanity classifier.\footnote{\url{https://github.com/snguyenthanh/better_profanity}} 
We acknowledge that automatic detection of profanity, toxicity, and hate speech are active research areas that have yet to be solved~\citep{dale-etal-2021-text, vidgen2021lftw}. 
Finally, for each country, we manually inspect 100 random posts to ensure syntactic and semantic correctness and find the remaining toxic or hate speech posts. We find that around 99\% of the posts are of good quality.

\paragraph{Data Statistics.} 
The final data statistics are shown in \Cref{tab:data_stats}.

\begin{table}[h]
\centering
\small
\setlength{\tabcolsep}{0.7em} %
{\renewcommand{\arraystretch}{1}%
\scalebox{0.9}{
\begin{tabular}{l M{1.2cm} M{1.2cm} |M{1.2cm} M{1.2cm}}
\toprule
\multirow{2}{*}{} & \multicolumn{2}{c}{\sc \# non-inspiring {{\color{Red}\xmark}}} & \multicolumn{2}{c}{\sc \# inspiring {\color{Goldenrod}\faLightbulbO}} \\
\midrule
  & Annotated & Weakly-Labeled & Annotated & Weakly-Labeled\\
  \midrule
   \worldflag[length=0.5cm, width=0.3cm]{IN}& 100 & 900 & 100 & 900 \\
    \worldflag[length=0.5cm, width=0.3cm]{GB}& 100 & 900 & 100 & 900\\
\bottomrule
      \end{tabular}
      }
      }
    \caption{Final number of inspiring ({\color{Goldenrod}\faLightbulbO}) and non-inspiring ({\color{Red}\xmark}) posts across India and the UK.\vspace{-1.8em}}
    \label{tab:data_stats}
\end{table}

\subsection{{\color{Thistle}\faLaptop} LLM-Generated Inspiring Content}
We generate 2,000 inspiring posts with GPT-4,\footnote{\url{https://platform.openai.com/docs/guides/text-generation/chat-completions-api}} balanced across India and the UK. Our study can be conducted with any LLM. However, we chose GPT-4 because it is one of the largest LLMs available and has been shown to effectively emulate human texts.~\cite{achiam2023gpt}

\subsubsection{Prompt Design and Robustness}
GPT-4 takes as input a list of \textit{message} objects, and returns an inspiring Reddit post. 
We use \textit{messages}, which are more interactive and dynamic compared to the classical prompt style. Specifically, we use messages with three different roles: \textit{system}, \textit{user} or \textit{assistant}.\footnote{\url{https://help.openai.com/en/articles/7042661-chatgpt-api-transition-guide}}

The prompt is first formatted with a \textit{system} role, which sets the behavior of the model. This is followed by a conversation between the \textit{user} and \textit{assistant}, in a \textbf{few-shot prompting} fashion.  Prior work found that LLMs function better with few-shot prompts (i.e., instructions alongside example output) rather than using zero-shot prompts (with no examples).~\citep{brown2020language}

\vspace{-0.5em}
\paragraph{System Prompt.}
We find that we can obtain high-quality responses with additional context in our prompts.
Therefore, we instruct the model to be a Reddit user from either the UK or India. To ensure that the generated data is diverse and reliable, we collect five versions of \textit{system} prompts with different phrasing, with one example shown below. 

\begin{quote} 
\textit{Imagine you're a person from \{location\} and use Reddit regularly.}
\end{quote}

\paragraph{User-Assistant Prompts.}
We design two rounds of conversations between a \textit{user} and an \textit{assistant}, where the \textit{user} asks for a Reddit post or comment, and the \textit{assistant} responds to the request. 
We use \textbf{few-shot prompting} by providing the \textit{assistant} in the first round of conversation with an inspiring post, which is randomly extracted from the real posts annotated as inspiring by the majority of annotators. 
Finally, the answer to the second \textit{user} is automatically generated by the \textit{assistant} and used to collect the GPT-4 generated inspiring post. A \textit{user} message is shown below.

\begin{quote} 
\textit{Write a Reddit post or comment of maximum 100 tokens about what inspires you.}
\end{quote}

\paragraph{Quality Assurance.}
To ensure the quality of our generated data, we conduct sanity checks to review approximately 200 inspiring posts balanced across the UK and India. 
The posts are checked for cultural knowledge, factuality, semantic and syntax errors, and style. Based on the feedback, we find that the posts are semantically and syntactically accurate, possess cultural knowledge, and are often more complex than real posts.
More information is provided in the \Cref{quality}.

\begin{table*}[h] 
\centering
    \small
    {\renewcommand{\arraystretch}{0.6}%
    \scalebox{0.9}{
    \begin{tabular}{l|p{7cm}|p{7cm}}
        \toprule
     & Real Post {\color{Thistle}\faUsers} & GPT4-generated Post {\color{Thistle}\faLaptop}  \\
        \midrule
        \multirow{2}{*}{\worldflag[length=0.5cm, width=0.3cm]{IN}} & {\color{Goldenrod}\faLightbulbO}  \textit{The youngest freedom fighter martyr of the country, the brave boy who refused to take the British cops across the river and was shot to death. His name was Baji Rout and he was 12 at the time of his death.} & {\color{Goldenrod}\faLightbulbO}  \textit{Dr. APJ Abdul Kalam - His humble beginnings, insatiable thirst for knowledge, and absolute dedication to his country have been my greatest inspiration. It pushed me to work harder, dream big, and contribute to society.} \\ 
         & {\color{Red}\xmark} \textit{What is a common meal there? I've always thought Indian palates to be quite spicy and rich.} & {\color{Red}\xmark} N/A \\ 
        \midrule 
        \multirow{2}{*}{\worldflag[length=0.5cm, width=0.3cm]{GB}} & {\color{Goldenrod}\faLightbulbO} \textit{Dr. Helen Sharman. I'm very pleased to come from a country whose first astronaut isn't a man by default. Equality shouldn't be about women catching up, it should be about women being first 50\% of the time.} & {\color{Goldenrod}\faLightbulbO} \textit{Absolutely love Sir David Attenborough's documentaries. His passion and commitment to preserving the environment is truly inspiring in these challenging times.} \\ 
         & {\color{Red}\xmark} \textit{Mate all he does is play football and misses pens. Not that special} & {\color{Red}\xmark} N/A \\ 
         \bottomrule
    \end{tabular}
    }
    }
    \caption{Random samples of inspiring ({\color{Goldenrod}\faLightbulbO}) and non-inspiring ({\color{Red}\xmark}) posts from India and UK.\vspace{-1.8em}}
    \label{tab:posts-examples}
\end{table*}

\paragraph{Cost.}
We generated 2,000 posts for two cultures, for a total cost of around 50\$.
(0.03 per 1K input tokens and 0.06 per 1K output tokens).

\section{Cross-Cultural Inspiration Analysis across Real and LLM-generated Posts}
In line with previous work~\citep{jakesch2023human}, we also find during manual inspection that it is challenging to differentiate between LLM-generated texts and those written by humans (see \Cref{tab:posts-examples}).
Therefore, we perform computational linguistic analyses to compare real and LLM-generated text across cultures.
This section addresses our first two research questions:
\textit{RQ1: How do inspiring posts compare across cultures?} and \textit{RQ2: How do AI-generated inspiring posts compare to real inspiring posts across cultures?}.

\subsection{Stylistic and Structural Features}
We assess the linguistic style and structure of real and LLM-generated inspiring posts across India and the UK in terms of (1) analytic writing, (2) descriptiveness, and (3) readability. 

\paragraph{Analytic Writing} index measures the complexity and sophistication of the writing, which can be an indicator of advanced thinking. 
The formula for analytic writing is $[articles + prepositions - pronouns - auxiliary verbs - adverb - conjunctions - negations]$ from LIWC scores~\citep{jordan2019examining, pennebaker2014small}
More information about LIWC can be found in \Cref{liwc}.
We display the scores in \Cref{tab:style}. The low complexity scores of Reddit posts are primarily negative due to their high usage of pronouns and lack of articles. We find that \textit{LLM-generated inspiring posts from the UK and India are more complex than real posts}, which aligns with our initial observations from data quality checks. At the same time, there is \textit{no significant difference in text complexity between real inspiring posts from India and those from the UK}.

\paragraph{Descriptiveness} can be measured by the frequency of adjectives used in language patterns. Texts with high rates of adjectives tend to be more elaborate and narrative-like compared to texts with low rates of adjectives.~\citep{chung2008revealing} In \Cref{tab:style}, we find that \textit{LLM-generated inspiring posts from the UK and India are more descriptive than real posts}. Additionally, \textit{real inspiring posts from India are more descriptive than those from the UK}.

\paragraph{Readability} considers not only word count, but also word complexity. For instance, longer words are more complex than shorter ones. We use the Flesch Reading Ease metric~\citep{flesch1948new}, which counts the number of words per sentence and syllables per word.
In \Cref{tab:style}, we find that \textit{LLM-generated inspiring posts are less readable than real posts}. Additionally, \textit{real inspiring posts from India are less readable than those from the UK}.
Furthermore, when measuring post length, \textit{LLM-generated UK posts are longer than real posts} and \textit{UK posts are shorter than Indian posts}.

\begin{table}
\centering
\setlength{\tabcolsep}{0.00em} %
{\renewcommand{\arraystretch}{1.3}%
\scalebox{0.7}{
\begin{tabular}{@{\extracolsep{7pt}}l c c c c}
  \toprule 
     &
    \multicolumn{2}{c}{\worldflag[length=0.5cm, width=0.3cm]{IN}} &
      \multicolumn{2}{c}{\worldflag[length=0.5cm, width=0.3cm]{GB}} 
      \\
      \cline{2-3} \cline{4-5} 
  \multicolumn{1}{l}{} & {\color{Goldenrod}\faLightbulbO} {\color{Thistle}\faUsers} &  {\color{Goldenrod}\faLightbulbO} \color{Thistle}\faLaptop & {\color{Goldenrod}\faLightbulbO} {\color{Thistle}\faUsers} & {\color{Goldenrod}\faLightbulbO} \color{Thistle}\faLaptop \\
  \midrule
Analytic & -16.7 $\pm$ 18.0  & -6.9 $\pm$ 10.0  & -17.1 $\pm$ 19.9 &-7.0 $\pm$ 11.4  \\
Descriptive & 8.2 $\pm$ 7.8  & 8.9 $\pm$ 3.6  & 6.6 $\pm$ 5.9 & 9.1 $\pm$ 3.9  \\
Readable & 36.3 $\pm$ 63.1  & 12.1 $\pm$ 20.6  & 53.2 $\pm$ 52.2 & 29.6 $\pm$ 18.0  \\
Word Count & 61.1 $\pm$ 59.8  & 66.9 $\pm$ 20.3  & 43.9 $\pm$ 49.3 & 49.9 $\pm$ 16.8   \\
\bottomrule
      \end{tabular}
      }
      }
    \caption{To what degree is the LLM-generated text ({\color{Thistle}\faLaptop}) stylistically and structurally different from the real text ({\color{Thistle}\faUsers})? 
    We compute the mean and standard deviation for the inspiring posts, across cultures. The differences are statistically significant, based on the Student t-test~\citep{student1908probable}, p-value < 0.05.\vspace{-1.8em}}
\label{tab:style}
\end{table}

\subsection{Semantic and Psycholinguistic Features}
We assess the semantic and psycholinguistic differences between LLM-generated and real inspiring posts across India and the UK, using topic modeling and LIWC psycholinguistic markers.

\paragraph{Data Pre-processing.}
We use \texttt{spaCy} library\footnote{\url{https://spacy.io/models}} to pre-process the data: tokenize each post, lowercase tokens, remove stop words, remove numbers, symbols, emojis, links, and lemmatize tokens.

\paragraph{Topic Modeling.}
We use Scattertext~\cite{kessler-2017-scattertext} to create interactive visualizations of linguistic patterns. 
We use the n-gram representation of the text data to conduct topic extraction through sentence-level clustering. 
\cite{ramos2003using}

\begin{figure*}
\centering
\includegraphics[width=0.9\linewidth]{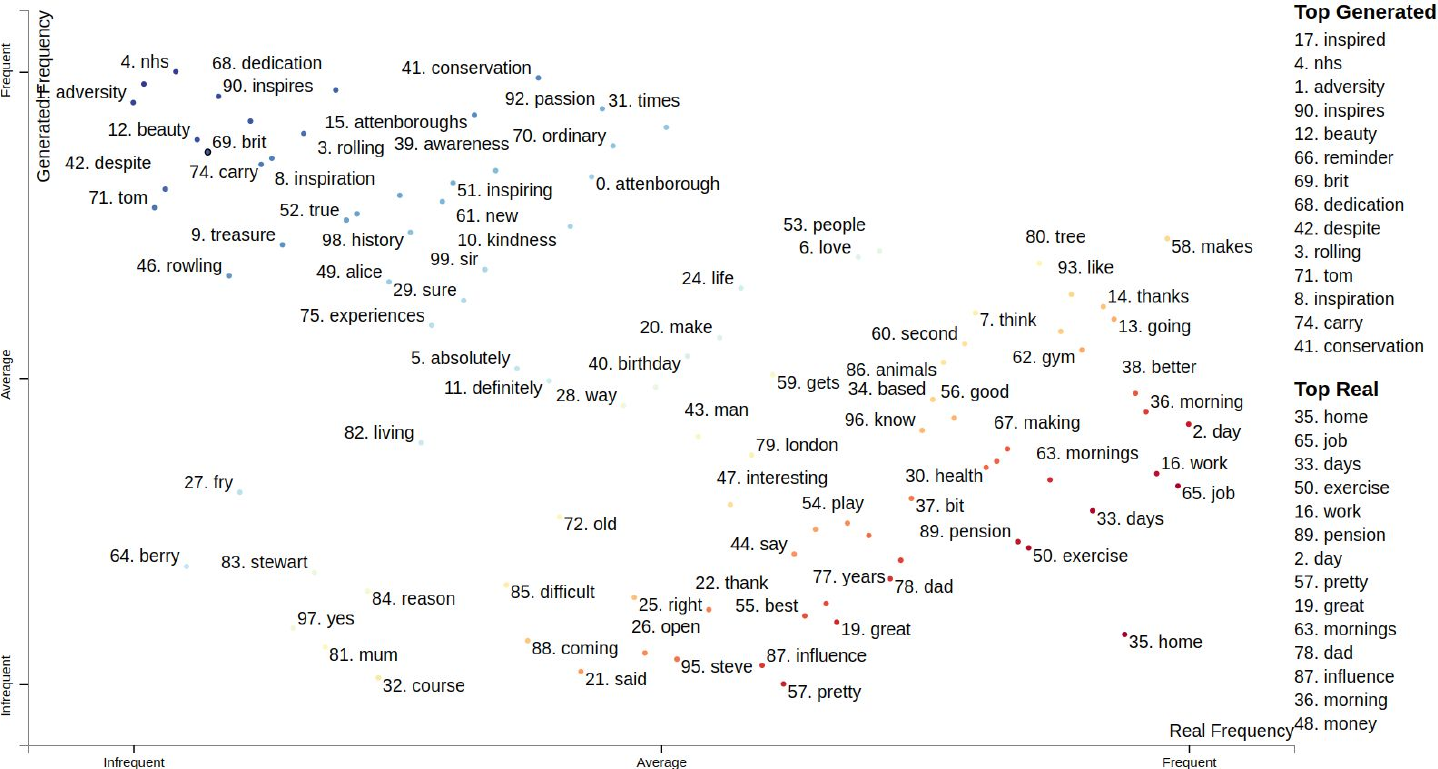}
\caption{Visualization of \textbf{topics} used in the real and generated ({\color{Thistle}\faUsers} vs. {\color{Thistle}\faLaptop}) inspiring posts from the \textbf{UK}. Points are colored {\color{Red}red} or {\color{Blue}blue} based on the association of their corresponding terms with UK Real inspiring posts or UK LLM-Generated inspiring posts. The most associated topics are listed under \textbf{Top Generated} and \textbf{Top Real} headings. Interactive version: \ttfamily\small\url{https://github.com/MichiganNLP/cross_inspiration}.\vspace{-1.8em}}
\label{fig:topic_gen_real_uk}
\end{figure*}

\textbf{Results.}  
We analyze the n-gram and topic distributions by various dimensions: inspiring vs. non-inspiring, Indian vs. UK, and real vs. generated.
We display the topic distribution across real and generated UK inspiring posts in \Cref{fig:topic_gen_real_uk}.
Further topic and n-gram distributions are in \Cref{topic-analysis}.

{\color{Goldenrod}\faLightbulbO} vs. {\color{Red}\xmark}: Comparing inspiring to non-inspiring real posts, we find that, in the Indian data (Fig. \ref{fig:conv_in}, \ref{fig:topic_in}), the topic of \textit{people} is frequently discussed in an inspiring context. 
Moreover, the most commonly occurring theme in non-inspiring posts is \textit{bot}, encompassing content regarding Reddit rules. 

In the UK data (Fig. \ref{fig:conv_uk}, \ref{fig:topic_uk}), the topic of \textit{life} is amongst the most commonly occurring themes in inspiring posts, with discussions surrounding \textit{career}, \textit{luck}, and \textit{pension}. On the other hand, the non-inspiring posts contain themes like \textit{dark}, including conversations about \textit{rain} and \textit{winter}.

{\color{Thistle}\faUsers} vs. {\color{Thistle}\faLaptop}: The LLM-generated Indian data (Fig. \ref{fig:conv_gen_real_in}, \ref{fig:topic_gen_real_in}) often places a significant emphasis on the topic of \textit{inspiring}, within which common words include \textit{dedication} and \textit{motivational}. There is also a significant number of posts surrounding \textit{isro (Indian Space Research Organisation)}, featuring terms such as \textit{space}, \textit{mars}, and \textit{mission}. 
In contrast, the real data from India contains mentions of \textit{housing}, \textit{summer}, and \textit{living}, grouped under the category \textit{live}. The \textit{movie} topic is also popular, containing words such as \textit{hype}, \textit{stardom}, and \textit{socialize}, hinting at the culture surrounding the film industry.

In the LLM-generated UK data (Fig. \ref{fig:topic_gen_real_uk}, top left), the topic \textit{nhs} is emphasized, where common words include \textit{pandemic}, \textit{heroes}, and \textit{staff}. Additionally, discussions often revolve around \textit{adversity}, with mentions of \textit{resilience}, \textit{determination}, and \textit{spirit}. 
In contrast, in the real data from the UK (Fig. \ref{fig:topic_gen_real_uk}, bottom right), discussions related to \textit{job} dominate, with mentions of \textit{salary}, \textit{savings}, and \textit{time}. Moreover, there is also a significant focus on \textit{exercise}, featuring words like \textit{discipline}, and \textit{motivate}.

\paragraph{LIWC Psycholinguistic Markers.} \label{liwc}
We use Linguistic Inquiry and Word Count (LIWC), a popular text analysis tool \citep{Pennebaker2007LinguisticIA,Pennebaker2015TheDA}, to obtain the words related to human cognitive processes from each post.
Specifically, we use the LIWC2015 dictionary, which contains 6,400 words and word stems, each related to a cognitive category. As an example, the word ``mother'' is assigned the following cognitive categories: \textit{female, family, social}. 
Even though it is less biased than embedding-based models, LIWC has limitations which we discuss in the Limitation section.

\textbf{Results.}
We display the top 24 categories and their LIWC scores, with the most significant differences across dimensions in \Cref{tab:liwc}. 
We find that \textit{male} words are more frequent than \textit{female}, the most common pronoun is \textit{I}, the most frequent tense is \textit{present}, and the most common emotion is \textit{positive}.

{\color{Goldenrod}\faLightbulbO} vs. {\color{Red}\xmark}:
Comparing inspiring to non-inspiring real posts, we find that the Indian inspiring posts are more related to \textit{socializing, insight, feelings, perception, affection}, and contain \textit{more positive emotions}.
Additionally, Indian inspiring posts tend to have \textit{more comparisons} and use more words related to \textit{achievement, health, reward}, and \textit{work}.

UK real inspiring posts contain more words related to \textit{affection, comparisons, feelings, achievement, health, home, leisure, money, reward}, and \textit{work} than non-inspiring posts.
Furthermore, compared to Indian inspiring posts, UK inspiring posts have fewer words related to \textit{family, socializing, affection, perception, religion}, \textit{less positive emotions} and more words related to \textit{achievement, health, home, leisure, money, rewards}, and \textit{work}.

{\color{Thistle}\faUsers} vs. {\color{Thistle}\faLaptop}:
Comparing real to LLM-generated inspiring posts, we find that real Indian posts tend to include more words related to \textit{family, social interactions, comparisons, feelings, perceptions} as well as \textit{home, leisure} and \textit{rewards}. 
Conversely, real Indian posts contain fewer words related to \textit{affection, insight, achievement, health}, and \textit{religion}.

UK real inspiring posts contain more words related to \textit{comparisons, feelings, health, home, leisure, money, rewards} and \textit{work} than LLM-generated UK posts.
Conversely, real UK posts contain fewer words related to \textit{socializing, affection, insight, perception, achievement} and \textit{religion} than LLM-generated UK posts.
Furthermore, compared to LLM-generated Indian posts, LLM-generated UK posts have fewer words related to \textit{socializing, achievement, leisure, money, religion, work} and more words related to \textit{affection, perception} and \textit{more positive emotions}.

\begin{table}
\centering
\setlength{\tabcolsep}{0.01em} %
{\renewcommand{\arraystretch}{1}%
\scalebox{0.60}{
\begin{tabular}{@{\extracolsep{7pt}}l c c c c c c}
  \toprule 
     &
    \multicolumn{3}{c}{\worldflag[length=0.5cm, width=0.3cm]{IN}} &
      \multicolumn{3}{c}{\worldflag[length=0.5cm, width=0.3cm]{GB}} 
      \\
      \cline{2-4} \cline{5-7} 
  \multicolumn{1}{l}{LIWC} & {\color{Goldenrod}\faLightbulbO} {\color{Thistle}\faUsers} & {\color{Red}\xmark} {\color{Thistle}\faUsers} & {\color{Goldenrod}\faLightbulbO} \color{Thistle}\faLaptop & {\color{Goldenrod}\faLightbulbO} {\color{Thistle}\faUsers} &  {\color{Red}\xmark} {\color{Thistle}\faUsers} & {\color{Goldenrod}\faLightbulbO} \color{Thistle}\faLaptop \\
  \midrule
  \multicolumn{7}{c}{Social Processes} \\
  \midrule
FAMILY & 0.4  & 0.4  & 0.2 & 0.2 & 0.1 & 0.2 \\
FRIEND & 0.3  & 0.4  & 0.2 & 0.2 & 0.4 & 0.2 \\
FEMALE & 0.4  & 0.3  & 0.2 & 0.4 & 0.4 & 0.2 \\
MALE & 1.1  & 1.0  & 0.9 & 0.9 & 1.3 & 1.7 \\
SOCIAL & 9.3  & 7.7  & 8.6 & 7.4 & 8.1 & 8.1 \\
  \midrule
  \multicolumn{7}{c}{Affective Processes} \\
  \midrule
AFFECT & 6.1  & 4.3  & 8.4 & 5.4 & 4.8 & 9.4 \\
NEGEMO & 1.4  & 1.3  & 1.0 & 1.3 & 1.6 & 0.9 \\
POSEMO & 4.5  & 3.0  & 7.3 & 4.1 & 3.1 & 8.3 \\
  \midrule
  \multicolumn{7}{c}{Cognitive Processes} \\
  \midrule
COMPARE & 3.0  & 2.4  & 1.8 & 3.0 & 2.7 & 1.8 \\
INSIGHT & 2.3  & 2.0  & 4.8 & 2.2 & 2.1 & 4.9 \\
  \midrule
  \multicolumn{7}{c}{Perceptual Processes} \\
  \midrule
FEEL & 0.6  & 0.4  & 0.4 & 0.7 & 0.3 & 0.4 \\
PERCEPT & 2.4  & 2.0  & 2.1 & 2.1 & 2.1 & 2.4 \\
  \midrule
  \multicolumn{7}{c}{Personal Concerns} \\
  \midrule
ACHIEV & 1.9  & 1.0  & 3.5 & 2.4 & 1.6 & 2.6 \\
HEALTH & 0.7  & 0.4  & 0.9 & 1.0 & 0.8 & 0.8 \\
HOME & 0.3  & 0.2  & 0.1 & 0.7 & 0.4 & 0.2 \\
LEISURE & 1.5  & 1.4  & 1.2 & 1.7 & 1.4 & 0.9 \\
MONEY & 0.7  & 0.8  & 0.6 & 1.2 & 0.8 & 0.4 \\
RELIG & 0.4  & 0.5  & 1.1 & 0.1 & 0.1 & 0.6 \\
REWARD & 2.0  & 1.1  & 1.0 & 2.2 & 1.6 & 1.1 \\
WORK & 3.0  & 1.9  & 3.0 & 3.9 & 2.7 & 1.9 \\
\bottomrule
      \end{tabular}
      }
      }
        \caption{Comparing LIWC scores across culture (India and the UK) in inspiring vs. non-inspiring ({\color{Goldenrod}\faLightbulbO} vs. {\color{Red}\xmark}) and real vs. LLM-generated posts ({\color{Thistle}\faUsers} vs. {\color{Thistle}\faLaptop}).
        \vspace{-1.5em}}
    \label{tab:liwc}
\end{table}

\section{Cross-Cultural Inspiration Detection across Real and LLM-generated posts} \label{sec:inspiration-detection}

To answer our last research question -- \textit{RQ3: Can detection models effectively differentiate inspiring posts across diverse cultures and data sources?} -- we fine-tune several classification models to identify if a post is inspiring, represents India or the UK, and whether it is real or LLM-generated. 
\paragraph{Implementation Details.}
As a baseline, we use a Random Forest classifier~\citep{ho1995random} with the default settings from \texttt{sklearn}~\citep{scikit-learn} and TF-IDF~\cite{ramos2003using} as features.
 
As a main model, we use the XLM-RoBERTa base model~\cite{roberta}, pre-trained on 2.5TB of filtered CommonCrawl~\cite{wenzek-etal-2020-ccnet}. We fine-tune the model on our dataset for five epochs with a learning rate of $2e-5$ and a batch size of $8$.
Finally, we also use the pre-trained Llama 2.7b model, from \texttt{HuggingFace}.\footnote{https://huggingface.co/meta-llama/Llama-2-7b} We fine-tune the model on our dataset using LoRA~\cite{hu2022lora} for five epochs with a rank of $4$, learning rate of $2e-4$ and a batch size of $2$.\footnote{All the model parameters are shared in the repository.}

\begin{figure}
\centering
\includegraphics[width=0.9\linewidth]{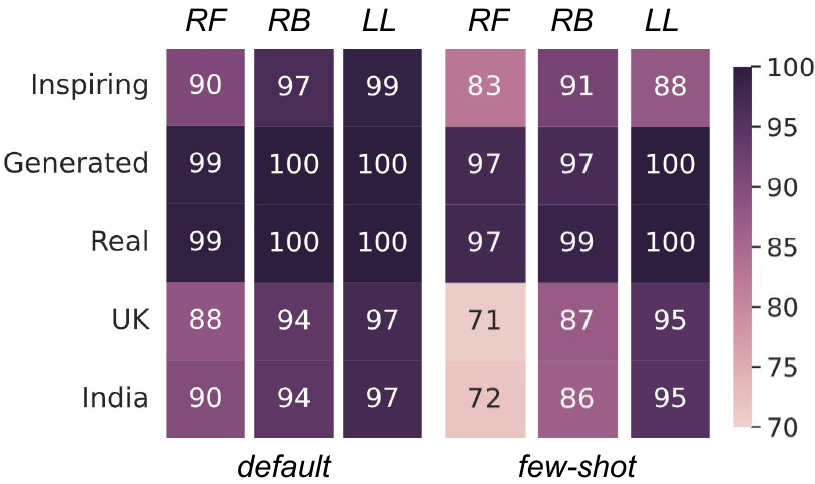}
\caption{Classification test accuracy with the \textit{few-shot} and \textit{default} setups with the Random Forest TF-IDF (\textit{RF}), XLM-RoBERTa base (\textit{RB}), and Llama 2.7b (\textit{LL}) models.
\vspace{-2em}}
\label{fig:both_eval}
\end{figure}

\paragraph{Model Training Setup.}
We experiment with two setups to split the data into training, validation, and test sets: a \textit{default} train-val-test split of 64-16-20\% and a \textit{few-shot} split of 8-2-90\%.
\paragraph{Results.}
Since the dataset is evenly distributed across classes, a random baseline results in a 50\% accuracy score. 
\Cref{fig:both_eval} displays the disaggregated results for each label. The Random Forest baseline with TF-IDF scores, trained in a \textit{default} setup achieves high accuracy scores ($88 - 99$), suggesting that important words and topics contribute significantly to inspiration detection across culture and source. This finding is also supported by the topic modeling results in Section 4.2, which shows different topic distributions for each class. Given that Random Forest is an interpretable model, we encourage future work to further analyse the importance of different features for inspiration detection. 
Finally, we find that even with very few training data (600 posts), in the \textit{few-shot} setup, the main models, Llama 2.7b and XLM-RoBERTa, learn to accurately distinguish inspiring content across cultures (India, UK) and data sources (real and generated), without significant differences in performance across cultures and data sources.

\section{Conclusion}
In this paper, we introduced the task of cross-cultural inspiration detection and generation in social media data. 
To facilitate research in this domain, we released \textsc{InspAIred}, a dataset of 2,000 real inspiring posts, 2,000 real non-inspiring posts, and 2,000 LLM-generated inspiring posts, evenly distributed across India and the UK. 
We performed extensive linguistic data analyses to gain insight into what topics inspire each culture and compare AI-generated inspiring posts to real inspiring posts across various linguistic dimensions. 
Despite the difficulty humans have in distinguishing between real posts and those generated by LLMs, we found that these posts have noticeable differences in style, structure, and semantics and that, even with little data, fine-tuned models accurately distinguish inspiring content across cultures and data sources.

We hope our work will enable the exploration of various applications to improve creativity and motivation, including storytelling, advertising, social media, therapy, and coaching.   
Our dataset is available to test, fine-tune, and analyze other models, and it can be accessed alongside our classification models at\url{https://github.com/MichiganNLP/cross_inspiration}.

\section*{Limitations}

\subsection*{A more fine-grained data split.}
Our posts are divided into two main categories - inspiring posts from India and inspiring posts from the UK. Initially, we tried to create a more detailed classification at the region or city level, but we faced difficulties in finding annotators from those specific regions. However, we encourage future research to explore this direction and investigate how inspiration varies within a country or region, as well as explore other demographic information, such as language, age, gender, or income.
\paragraph{Study case for India.}
The extensive discussions on Indian subreddits, like r/AskIndia and r/India, offer straightforward access to text-based motivational narratives. Additionally, region-based subReddits, like r/Chennai and r/Mumbai, provide diverse insights into localized experiences and discussions.
The nature of posts on this topic is mostly anecdotal. Reddit users contribute by sharing personal experiences that have inspired them. 

\subsection*{A limited number of cross-cultural inspiring posts.}
It is crucial to recognize that inspiration is not always explicitly articulated through words like ``inspiring'' or ``motivation'', especially within cultures that aren't as open about talking about such topics. In many instances, it manifests more subtly, embedded within narratives, imagery, or cultural expressions. This implicit nature of inspiration adds another layer of complexity to the data collection process. 

The language barrier also introduces an additional barrier, as it requires linguistic fluency and cultural understanding to interpret and analyze inspiring content effectively. Posts in languages other than English may contain nuances and cultural references that are not easily translatable. 
Moreover, it becomes more challenging to collect inspiring content when data collection is restricted to certain countries. 
That is why we collected fewer posts compared to Ignat et al.'s ``general'' inspiring posts.

\subsection*{Data generated with a closed-source model.}
We use GPT-4 to generate the inspiring posts, which is not an open-source LLM. We believe it would be valuable for future work to consider generating more data using open-source models like LLaMA or BLOOM.
At the same time, we publicly released the data generated with GPT-4 so that others can build on this dataset, and chose this model due to its SOTA performance, worldwide accessibility, and popularity.

\subsection*{Relevance of LLM-based data to current times.}

In our reliance on LLM-based data, it's imperative to recognize the temporal limitations inherent in its training corpus. While these models offer remarkable capabilities in understanding and generating text, they might not fully capture the current cultural and societal context in their outputs. This limitation can be primarily attributed to the fact that these models are trained on data only up to a specific cutoff date, i.e., GPT-4 only learned from data dated up to January 2022.

\subsection*{Limitations of the LIWC-based data analysis.}
LIWC is a dictionary where each word is associated with a psycholinguistics category. Therefore, it is less biased than embedding-based models that often contain biased word associations (e.g., man-programmer, woman-homemaker).
Based on a simple and scalable word count approach, LIWC is one of the most commonly used data analysis tools. However, it has several limitations.
First, it does not consider word sense disambiguation based on context, which can lead to inaccurate results~\cite{schwartz2013choosing}. 
Second, the model was developed and validated for the
analysis of long-form writing, which may impact its validity on short social media posts~\cite{Panger2016ReassessingTF}. 
Third, ongoing research is being conducted to measure the extent to which LIWC demonstrates cross-linguistic and cross-cultural generalizability, given that most studies have been conducted in English~\cite{Dudu2021PerformingMA}.

\bibliography{main.bib}

\appendix

\section{Appendix}

\subsection{Data Quality Assurance} \label{quality}

We mention in the paper that the posts are checked for cultural knowledge, factuality, semantic and syntax errors, and style. 
Specifically, the authors each check 200 real and generated posts from each culture.
One author is from India and is, therefore, able to authenticate the cultural and factual aspects of the sampled posts (e.g., Sal Khan and his education contributions, the ISRO, and its scientific progress in India) and check the correctness of the few (<10\% of the data) code switch posts between English and Tamil.
All the authors possess cultural knowledge about the UK, to recognize the cultural and factual elements of the posts. When unsure, they manually searched Wikipedia for the factuality of the posts (e.g., Sir David Attenborough's contributions to nature, the British Nicola Adams, the first-ever female boxer to retain an Olympic title).

Finally, all the authors possess proficient English skills to check for semantic and syntax errors in the sampled posts. 
We find that apart from a few typos (only in real posts), very common in social media posts, the vast majority of the posts (98\%) are correct.
In terms of style, we observe that LLM-generated sampled posts are more formal and complex than real posts (e.g., there are no typos, and there are more articles and prepositions). Our analysis also confirmed this by automatically measuring the analytic index.

\subsection{Fine-tuning Process} \label{finetune}

\paragraph{XLM-RoBERTa Model.} 

\textbf{Implementation Details.}
We experiment with three different fine-tuned classification models for the purpose of weakly labeling the presence of inspiration in the dataset. 
The first attempt involved fine-tuning the model on the general inspiration dataset introduced in~\cite{Ignat2021DetectingIC}. 
In each of the experiments, the data is split into three subsets: training, validation, and test sets, using an 80:10:10 ratio. 
The fine-tuning process involved training the model for 5 epochs with a learning rate of $2e-5$ and a batch size of $8$. We monitored the model performance on a separate validation set and selected the best model checkpoint based on accuracy metric. 

In the second experiment, we focused on a subset of the dataset used in annotation, consisting of 200 posts. These posts were labeled as inspiring if at least one user considered them to be so. We fine-tuned the base XLM-RoBERTa model using this subset, maintaining the same training configurations as in the initial experiment. In the last experiment, we leveraged the user-annotated posts to further refine the model trained on the general inspiration dataset from the first experiment. 
This approach aimed to leverage the insights gained from the initial fine-tuning process and apply them to the user-annotated subset, thereby refining the model to better capture the nuances present in this specific dataset.

\subsection{Topic Analysis} \label{topic-analysis}

\begin{figure*}
\centering
\includegraphics[width=\linewidth]{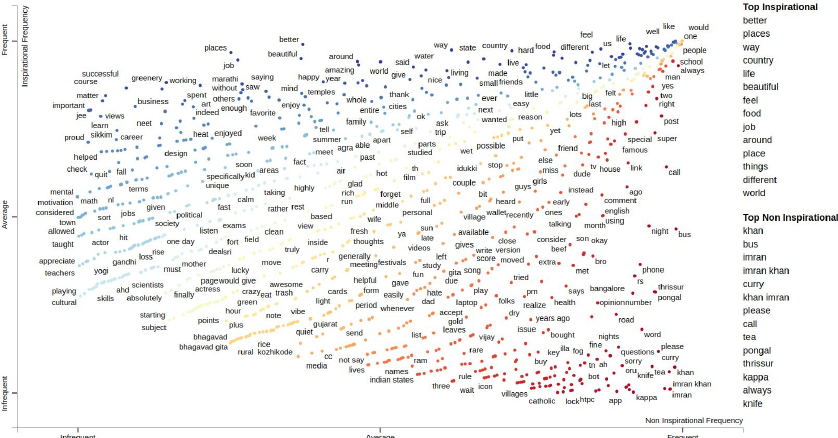}
\caption{Scattertext visualization of \textbf{unigrams} used in the real inspiring and non-inspiring ({\color{Goldenrod}\faLightbulbO} vs. {\color{Red}\xmark}) Reddit posts from \textbf{India}. Points are colored in {\color{Red}red} or {\color{Blue}blue} based on the association of their corresponding terms with Indian Non-inspiring posts or Indian inspiring posts. The most associated terms are listed under ``Top inspiring'' and ``Top Non-inspiring'' headings.}
\label{fig:conv_in}
\end{figure*}

\begin{figure*}
\centering
\includegraphics[width=\linewidth]{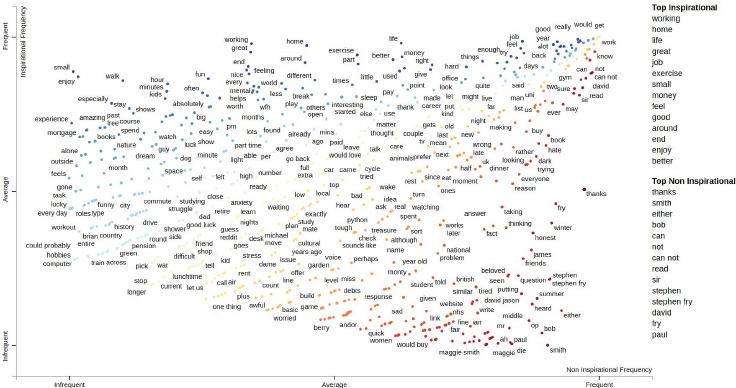}
\caption{Scattertext visualization of \textbf{unigrams} used in the real inspiring and non-inspiring ({\color{Goldenrod}\faLightbulbO} vs. {\color{Red}\xmark}) Reddit posts from the \textbf{UK}. Points are colored in {\color{Red}red} or {\color{Blue}blue} based on the association of their corresponding terms with UK Non-inspiring posts or UK inspiring posts.
The most associated terms are listed under ``Top inspiring'' and ``Top Non-inspiring'' headings. }
\label{fig:conv_uk}
\end{figure*}

\begin{figure*}
\centering
\includegraphics[width=\linewidth]{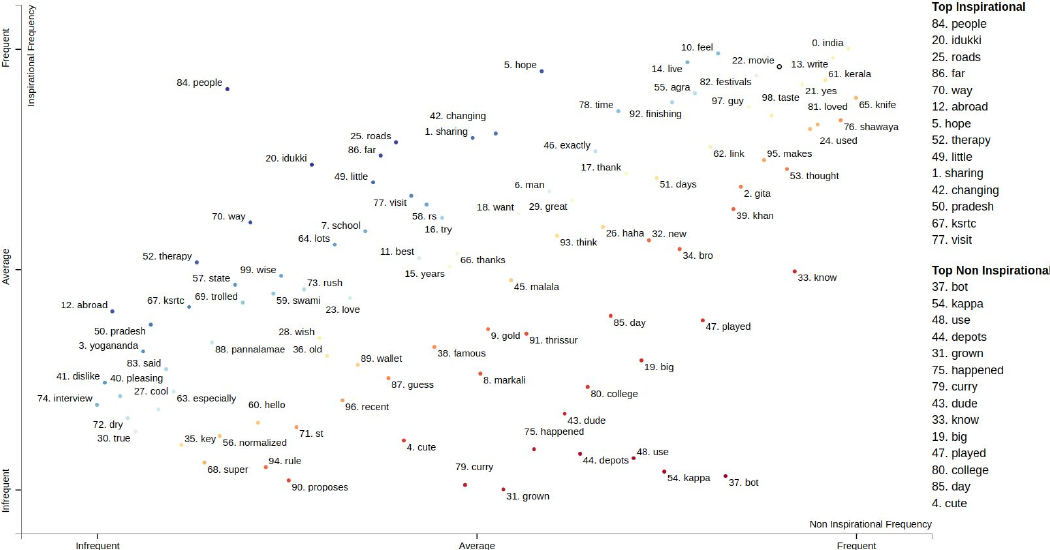}
\caption{Scattertext visualization of \textbf{topics} used in the real inspiring and non-inspiring ({\color{Goldenrod}\faLightbulbO} vs. {\color{Red}\xmark}) Reddit posts from \textbf{India}. Points are colored in {\color{Red}red} or {\color{Blue}blue} based on the association of their corresponding terms with India Non-inspiring posts or India inspiring posts. The most associated topics are listed under ``Top inspiring'' and ``Top Non-inspiring'' headings. }
\label{fig:topic_in}
\end{figure*}

\begin{figure*}
\centering
\includegraphics[width=\linewidth]{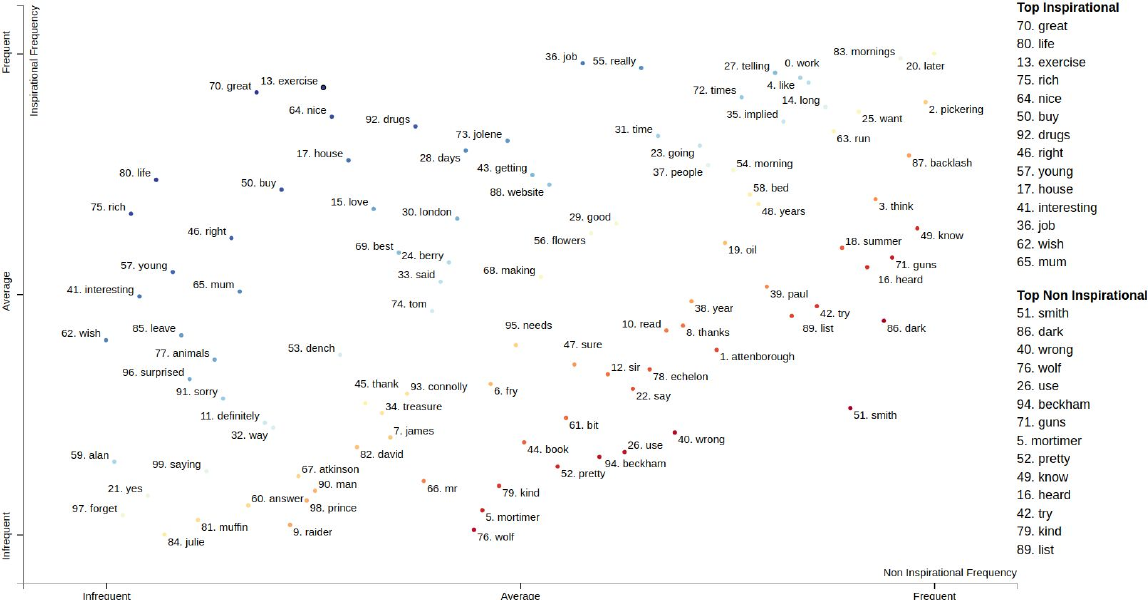}
\caption{Scattertext visualization of \textbf{topics} used in the real inspiring and non-inspiring ({\color{Goldenrod}\faLightbulbO} vs. {\color{Red}\xmark}) Reddit posts from the \textbf{UK}. Points are colored {\color{Red}red} or {\color{Blue}blue} based on the association of their corresponding terms with UK Non-inspiring posts or UK inspiring posts. 
The most associated topics are listed under ``Top inspiring'' and ``Top Non-inspiring'' headings. }
\label{fig:topic_uk}
\end{figure*}

\begin{figure*}
\centering
\includegraphics[width=\linewidth]{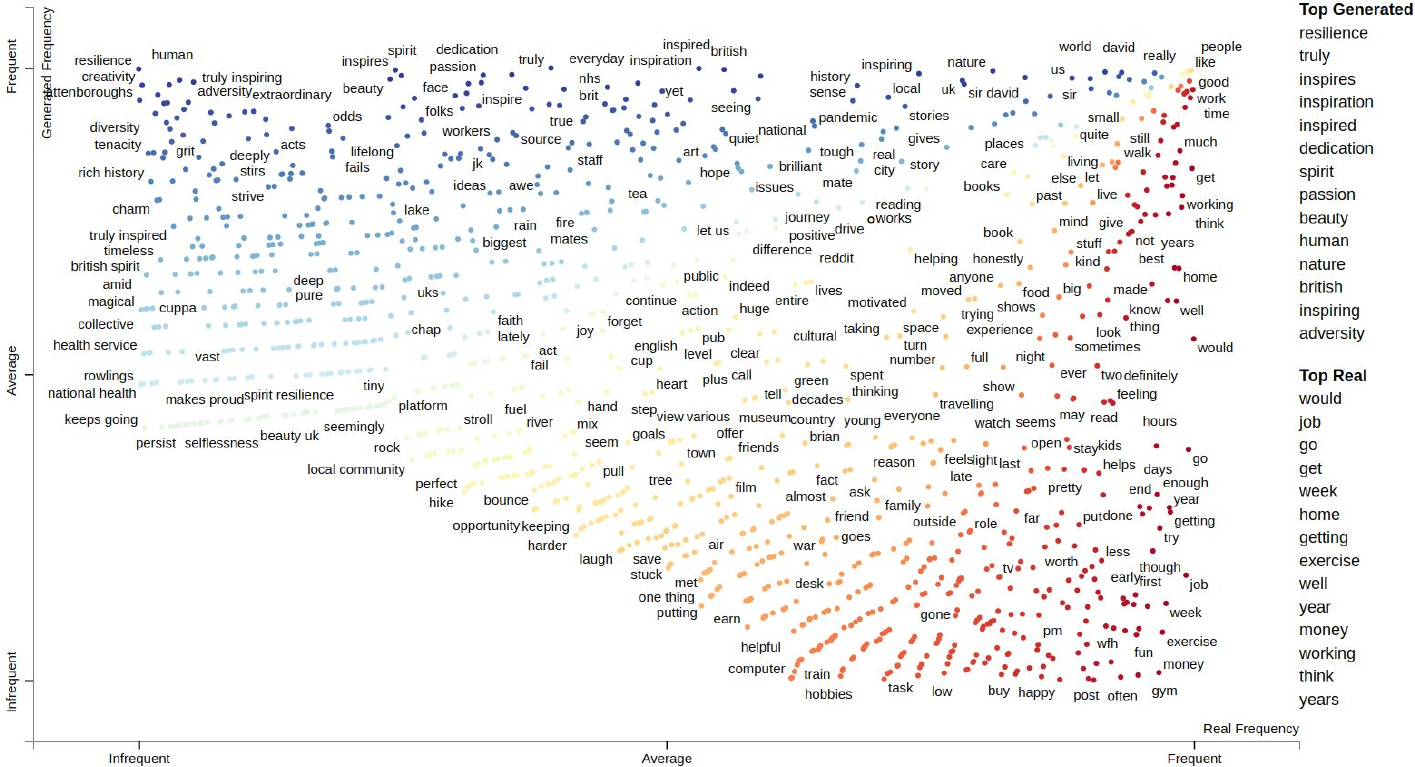}
\caption{Scattertext visualization of \textbf{unigrams} used in the real and generated ({\color{Thistle}\faUsers} vs. {\color{Thistle}\faLaptop}) inspiring posts from the \textbf{UK}. Points are colored {\color{Red}red} or {\color{Blue}blue} based on the association of their corresponding terms with the UK Real inspiring posts or the UK LLM-Generated inspiring posts. The most associated topics are listed under \textbf{Top Generated} and \textbf{Top Real} headings. }
\label{fig:conv_gen_real_uk}
\end{figure*}

\begin{figure*}
\centering
\includegraphics[width=\linewidth]{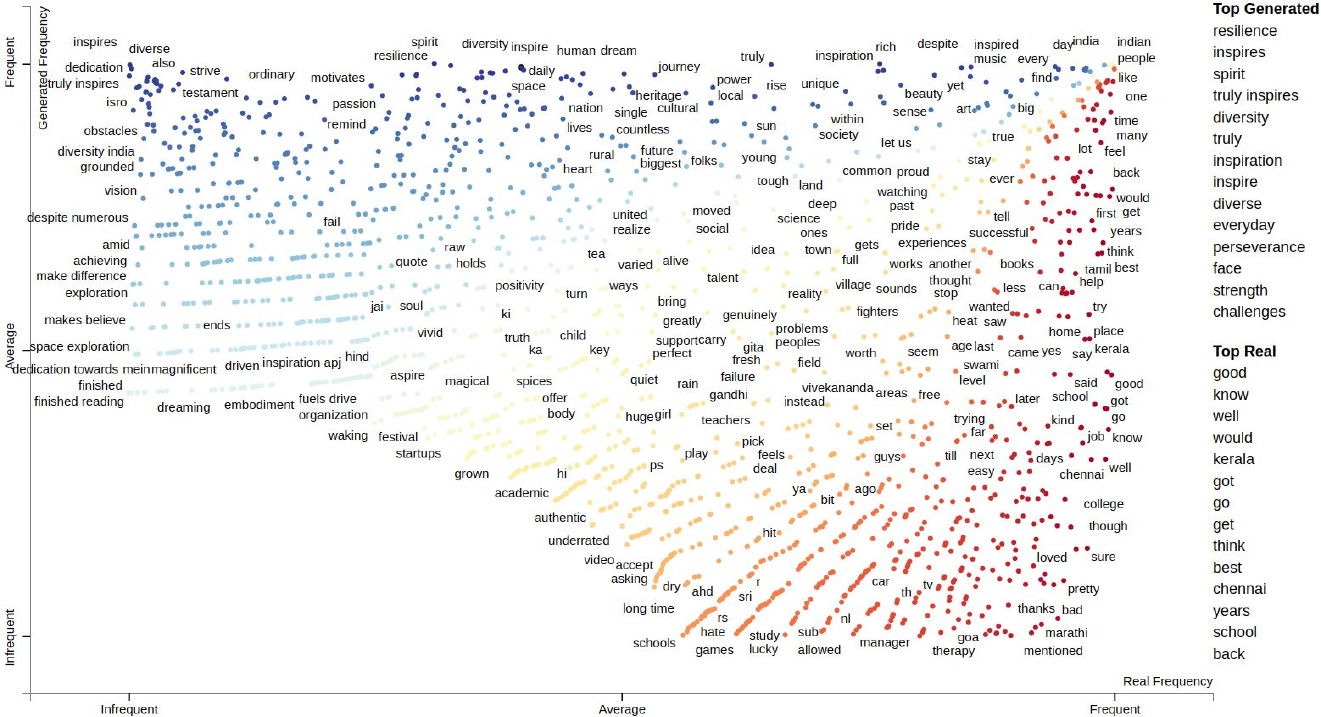}
\caption{Scattertext visualization of \textbf{unigrams} used in the real and generated ({\color{Thistle}\faUsers} vs. {\color{Thistle}\faLaptop}) inspiring posts from \textbf{India}. Points are colored {\color{Red}red} or {\color{Blue}blue} based on the association of their corresponding terms with India Real inspiring posts or India LLM-Generated inspiring posts. The most associated topics are listed under \textbf{Top Generated} and \textbf{Top Real} headings. }
\label{fig:conv_gen_real_in}
\end{figure*}

\begin{figure*}
\centering
\includegraphics[width=\linewidth]{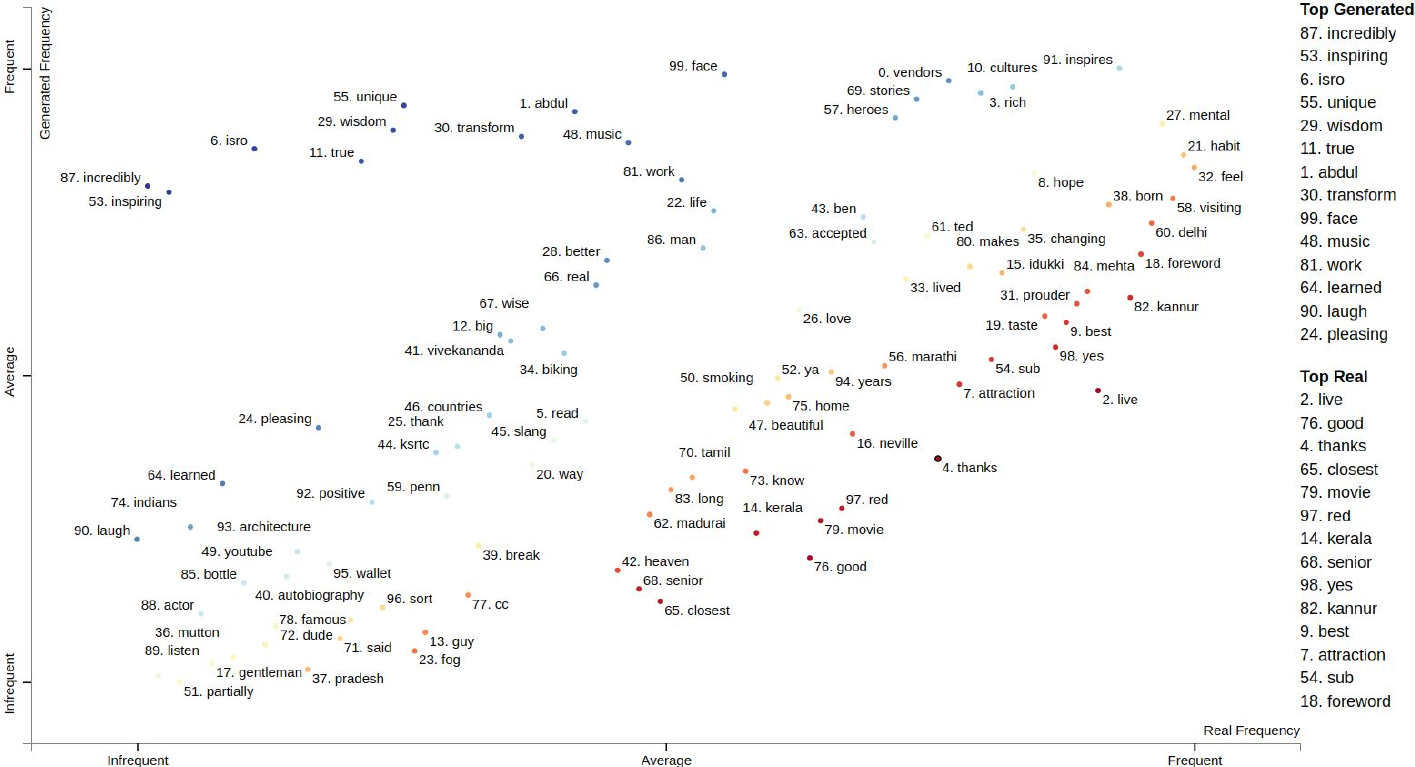}
\caption{Scattertext visualization of \textbf{topics} used in the real and generated ({\color{Thistle}\faUsers} vs. {\color{Thistle}\faLaptop}) inspiring posts from \textbf{India}. Points are colored {\color{Red}red} or {\color{Blue}blue} based on the association of their corresponding terms with India Real inspiring posts or India LLM-Generated inspiring posts. The most associated topics are listed under \textbf{Top Generated} and \textbf{Top Real} headings. }
\label{fig:topic_gen_real_in}
\end{figure*}

\end{document}